\documentclass[letterpaper, 10 pt, conference]{ieeeconf}  

\IEEEoverridecommandlockouts                              

\usepackage[T1]{fontenc}
\usepackage{graphics} 
\usepackage{epsfig} 
\usepackage{mathptmx} 
\usepackage{times} 
\usepackage{amsmath} 
\usepackage{amssymb}  
\usepackage{booktabs,threeparttable,makecell,array}

\usepackage{multirow}
\usepackage[table,xcdraw]{xcolor}
\usepackage{threeparttable}
\usepackage{eucal}
\usepackage{balance}
\usepackage{mathrsfs}  
\usepackage{comment}
\usepackage[noend]{algpseudocode}
\usepackage{algorithm}
\usepackage{float}
\usepackage{textcomp}
\usepackage{mathcomp}
\usepackage{bm}
\usepackage{cite} 
\usepackage{booktabs}
\usepackage{url}
\usepackage{hyperref}
\usepackage{comment}
\usepackage[whole]{bxcjkjatype} 
\usepackage{threeparttable} 
\usepackage{subcaption}
\usepackage{afterpage}
\usepackage[section]{placeins} 

\hypersetup{
    pdfauthor={},
    pdftitle={VLG-Loc: Vision-Language Global Localization from Labeled Footprint Maps}, 
    pdfcreator={}, 
    pdfproducer={}
}

\usepackage{array}
\def\eg{{\it e.g.}}

\def\ie{{\it i.e.}}

\newcommand{\mytag}[1]{\texttt{#1}}

\title{\LARGE \bf
VLG-Loc: Vision-Language Global Localization\\from Labeled Footprint Maps
}
\author{
Mizuho Aoki$^{1}$\thanks{*Work done during an internship at CyberAgent.}, %
Kohei Honda$^{1,2}$, Yasuhiro Yoshimura$^{2}$,Takeshi Ishita$^{2}$, Ryo Yonetani$^{2}$
\thanks{$^{1}$The Department of Mechanical Systems Engineering, Nagoya University, Aichi, Japan, {\tt\small\ mizuhoaoki1998@gmail.com}
}%
\thanks{$^{2}$CyberAgent AI Lab, Tokyo, Japan, {\tt\small\{honda\_kohei, ishita\_takeshi, yoshimura\_yasuhiro, yonetani\_ryo\}@cyberagent.co.jp}
}%
}

\begin{document}

\onecolumn

This work has been submitted to the IEEE for possible publication. Copyright may be transferred without notice, after which this version may no longer be accessible.

\cleardoublepage
\twocolumn

\maketitle
\thispagestyle{empty}
\pagestyle{empty}


\newcommand{\red}[1]{\textcolor{red}{#1}}
\newcommand{\aoki}[1]{\textcolor{blue}{#1}}
\newcommand{\yonetani}[1]{\textcolor{green}{#1}}
\newcommand{\honda}[1]{\textcolor{yellow}{#1}}
\newcommand{\yoshimura}[1]{\textcolor{orange}{#1}}
\newcommand{\ishita}[1]{\textcolor{purple}{#1}}


\begin{abstract}
This paper presents 
Vision-Language Global Localization (VLG-Loc), a novel global localization method that uses human-readable labeled footprint maps containing only names and areas of distinctive visual landmarks in an environment.
While humans naturally localize themselves using such maps, translating this capability to robotic systems remains highly challenging due to the difficulty of establishing correspondences between observed landmarks and those in the map without geometric and appearance details.
To address this challenge, VLG-Loc leverages a vision-language model (VLM) to search the robot's multi-directional image observations for the landmarks noted in the map. The method then identifies robot poses within a Monte Carlo localization framework, where the found landmarks are used to evaluate the likelihood of each pose hypothesis. 
Experimental validation in simulated and real-world retail environments demonstrates superior robustness compared to existing scan-based methods, particularly under environmental changes. Further improvements are achieved through the probabilistic fusion of visual and scan-based localization.
\end{abstract}

\section{Introduction}

Imagine you get lost in a large supermarket. A map on the wall shows only the locations and names of the sections. In front of you is the snack aisle, and to your left is the drinks corner. By finding these sections on the map, you can pinpoint your location. This scenario demonstrates how we humans perform \emph{global localization} in cluttered environments. Many of the maps we see every day just describe landmarks (\eg, section names or other facilities like entrances for retail stores) primarily with text. Consequently, we rely more on the visual identity of these landmarks than on their precise geometric features.

\begin{figure}[t]
\centering
\includegraphics[width=0.85\linewidth]{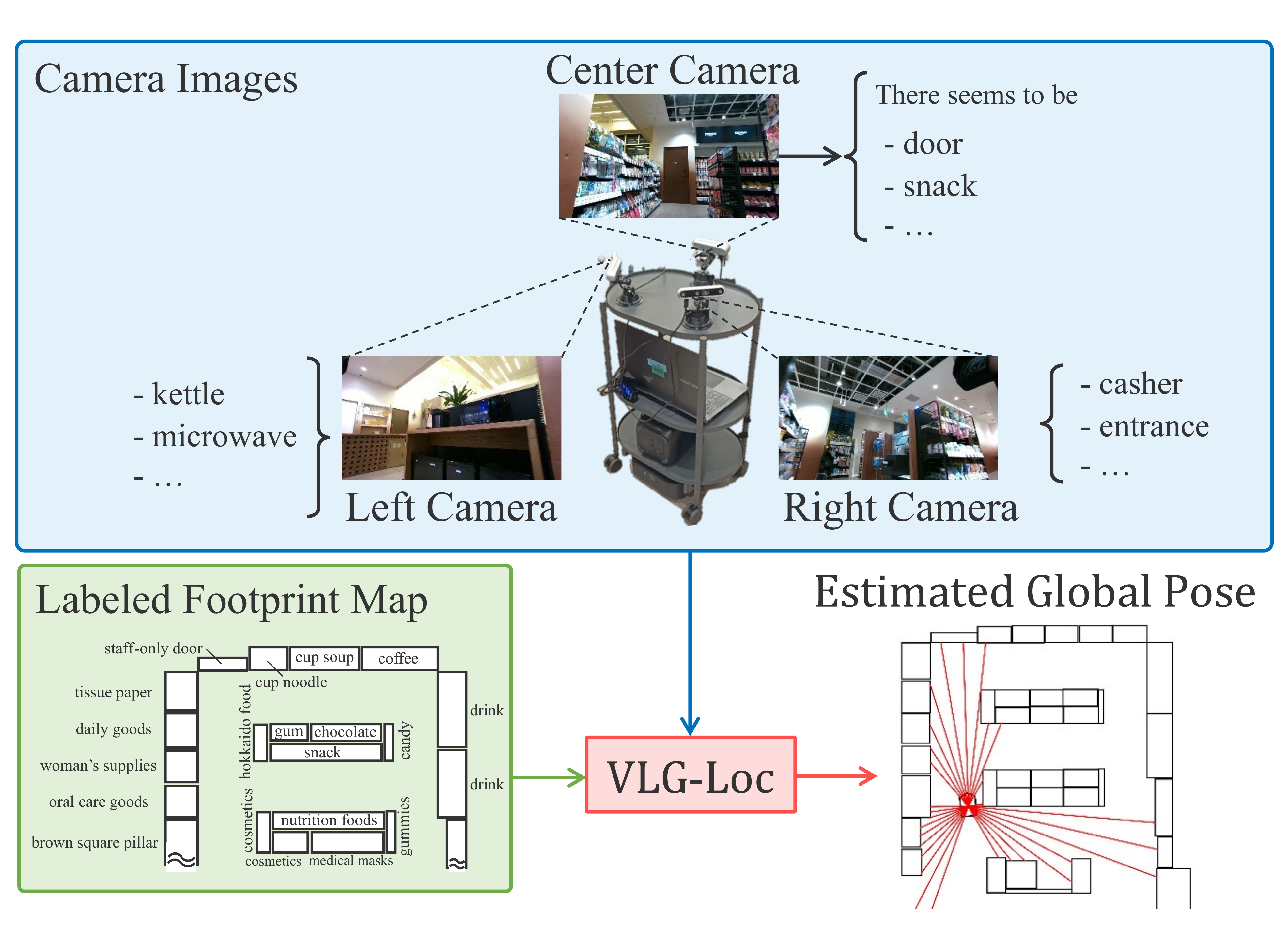}
\caption{Given camera images and a labled footprint map, VLG-Loc leverages a vision-language model to detect ``what landmarks are in the image'' and matches them to estimate the robot\textquotesingle s global pose.}
\label{fig:teaser}
\vspace{-6mm}
\end{figure}

Our goal is to endow robots with this type of global localization ability based on simplified maps. Unlike dense geometric map representations commonly used in robotics~\cite{lidar_global_loc_survey, vision_global_loc_survey}, we will use human-readable \emph{labeled footprint maps} that provide only the names and areas of a small number of distinctive landmarks (see Fig.~\ref{fig:teaser}). Given query image observations from a robot and the labeled footprint map of the environment, the task is to estimate the robot's pose (positions and orientations) by establishing correspondences between observed and mapped landmarks. 

Labeled footprint maps offer several compelling advantages and technical challenges: these maps are readily available in many structured environments (\eg, shopping malls, airports). They are also easy to update against environmental changes such as daily product rotations and seasonal furniture rearrangements, compared to conventional dense maps (\eg, point clouds) that need labor-intensive data collection. On the other hand, the lack of geometric or visual details in these maps makes landmark association particularly difficult. Existing methods typically rely on local descriptors of shapes and appearances of the surrounding obstacles~\cite{3dbbs2024icra, locnerf2023}, which are hard to extract from text-based map representations.

In this work, we propose a novel global localization method named \emph{Vision-Language Global Localization (VLG-Loc)}%
\footnote{The source code and dataset are available at: \url{https://github.com/CyberAgentAILab/VLG-Loc}}
At its core, VLG-Loc uses a labeled footprint map in two ways. (1) We employ a vision-language model (VLM) to search image observations for landmarks that exist in the map (\ie, \emph{visible} set of landmarks). (2) For a pose hypothesis, we also simulate landmarks that should be visible in the map (\ie, \emph{simulated} set of landmarks). Comparing these two sets allows us to evaluate the likelihood of the pose hypothesis in terms of landmark visibility. Maximum likelihood estimation within a standard Monte Carlo localization (MCL) framework~\cite{fox1999mcl} estimates the most likely pose for given image observations. This way, VLG-Loc does not require geometric or appearance descriptors to be extracted from observations and maps unlike most existing solutions. Interestingly, the proposed VLG-Loc performs complementarily to existing scan-based localization methods; VLG-Loc provides rough but unimodal pose estimates when unique landmarks are present, while scan-based localization tends to be more precise but potentially multimodal due to repetitive geometric structures. We will show that probabilistically fusing these two methods whenever possible further leads to more accurate and robust localization.

We evaluate the proposed VLG-Loc comprehensively in simulated and real-world retail environments. Experimental results demonstrate the superior robustness of VLG-Loc as well as its probabilistic fusion with scan-based localization compared to existing methods~\cite{fox1999mcl, cbgl2024iros}, particularly in the presence of environmental changes. 

\section{Related Work}

This section reviews existing methods that can be categorized into scan-based and vision-based approaches. We also introduce recent work that employs landmark-based sparse map representations.

\subsection{Scan-based Global Localization}

Scan-based localization employs 2D or 3D LiDAR sensors to match observations and point-cloud maps based on the geometric details of surrounding objects. Monte Carlo Localization (MCL)~\cite{fox1999mcl,probabilistic_robotics} is a representative framework, which estimates the most probable poses via Bayesian estimation with Monte Carlo approximation. Thanks to its ability to deal with non-parametric likelihood distributions, MCL is known to be robust to environmental uncertainty where the LiDAR observations match several locations in the environment.
Another approach relies on scan-to-map registration~\cite{freespace2019iros, icp, 2dndt}, which estimates alignments of point clouds between observations and maps for pose estimation. To achieve both estimation accuracy and efficiency, recent work has first selected promising pose hypotheses from a uniform prior over the map and then performed the local scan-to-match alignment~\cite{cbgl2024iros}.

The main limitation of scan-based methods is their sensitivity to geometrically repetitive or feature-sparse environments, such as wide uniform corridors and tunnels. In such cases, likelihoods can become ambiguous and registration may converge to incorrect solutions. The use of 3D LiDAR sensors would provide richer geometric cues for improved robustness~\cite{teaser2021tro, 3dbbs2024icra, single2022icra}, nevertheless in exchange for higher hardware and computation costs.

\subsection{Vision-based Global Localization}
Vision provides rich appearance cues that enable accurate global localization even in environments with limited geometric structure. 
A wide range of approaches has been proposed, including image retrieval~\cite{netvlad}, structure from motion~\cite{sfm2016cvpr}, and neural rendering~\cite{locnerf2023}, to demonstrate successful performance across diverse scenarios. Recent work has further attempted to integrate additional modalities~\cite{instant_lidar_visual}, text recognition using vision language models~\cite{niu2025llmloc}, or semantic reasoning~\cite{fmloc2023iros} to improve robustness.

Because vision-based approaches rely on the appearance-level similarity between observations and maps, they are inherently vulnerable to featureless or visually repetitive scenes~\cite{vision_global_loc_survey}. Dynamic changes or view dependency (\eg, mirrors) of appearances can also raise practical challenges and complicate localization pipelines.

\subsection{Landmark-based Sparse Map Representations}

Dense maps are crucial for both scan-based and vision-based approaches to enable robust matching, yet maintaining them is computationally and memory-intensive. To address this issue, recent work has explored sparse map representations by discovering distinct landmark objects from the environment. For example, \cite{clip_clique,clip_loc} detect objects from both query and map images and evaluates their semantic similarity through contrastive language-image pretraining (CLIP)-based encoding~\cite{radford2021learning}. 
\cite{lm_nav2022corl} proposed a navigation approach based on a topological map, where camera observations were linked by their relative spatial relations.
A concurrent work~\cite{paul2025sparseloc} has alternatively adopted semantic segmentation and VLM-based image captioning to enable open-set landmark discovery.

While our work and these methods share the idea of relying on sparse landmarks for global localization, the key difference is in how to construct and maintain map databases. Existing methods require camera-based or LiDAR-based sensing to provide visual and geometric information to describe landmarks. 
This incurs expensive labor costs for environments with frequent updates in their appearances, such as retail stores with regular product rotations. The use of labeled footprint maps in our method mitigates this limitation since map maintainers just have to update labels for such updates.
A related idea is the attempt to localize in large department stores by exploiting abundant textual signs such as shop names~\cite{lost_shopping_2015}.
In contrast, our method leverages the capability of vision–language models to associate abstract labels with visual observations.
This enables global localization in a wider range of environments.

\section{Preliminaries}

\subsection{2D Global Localization Problem}
Global localization is a problem of determining the global pose of a mobile robot within a known map of the environment from the robot's observations. We particularly address the 2D global localization problem where the map and robot's position are both represented in the 2D space $\Omega\subset\mathbb{R}^2$. Robot poses are given in the special Euclidean space, $\bm{x}=(x, y, \theta)\in \textbf{SE}(2)$, consisting of 2D positions $(x, y) \in\Omega$ and 1D rotations $\theta\in[-\pi, \pi)$ (\ie, yaw). This is a typical problem setup for on-ground mobile robots navigating planar environments~\cite{lidar_global_loc_survey}.

\subsection{Probabilistic Formulation and Monte Carlo Solution}
\label{sec:loc_formulation}
Let $\mathcal{M}, \mathcal{Z}$ be the map and the robot's observation, respectively. The process to infer the pose from those inputs is formulated probabilistically as follows:
\begin{align}
    p(\bm{x}\mid \mathcal{M}, \mathcal{Z})\propto p(\mathcal{Z}\mid \bm{x}, \mathcal{M})p(\bm{x}\mid\mathcal{M}).
\end{align}
The maximum a posteriori (MAP) estimate of the above is:
\begin{align}
\bm{x}^*
=
\underset{\bm{x}}{\arg\max}\;
p(\mathcal{Z} \mid \bm{x}, \mathcal{M})\, p(\bm{x} \mid \mathcal{M}).
\label{eq:loc_general_map}
\end{align}

The global localization assumes no information given to the prior $p(\bm{x}\mid\mathcal{M})$. Eq.~(\ref{eq:loc_general_map}) then reduces to the maximum likelihood estimation:
\begin{align}
\bm{x}^*
&=
\underset{\bm{x}}{\arg\max}\;
p(\mathcal{Z} \mid \bm{x}, \mathcal{M})
=
\underset{\bm{x}}{\arg\max}\;
\log p(\mathcal{Z} \mid \bm{x}, \mathcal{M}).
\label{eq:loc_mle}
\end{align}

A typical solution to this probabilistic formulation is Monte Carlo Localization (MCL)~\cite{fox1999mcl}, which approximates the distribution by a finite set of hypotheses sampled from the state space. Particularly for global localization problems, those hypotheses are sampled uniformly and weighted by the log-likelihood function $\log p(\mathcal{Z} \mid \bm{x}, \mathcal{M})$ with the observation $\mathcal{Z}$. The pose estimation $\bm{x}^*$ is given by the highest-weight hypothesis or the weighted mean of the hypotheses in the most dominant cluster.

MCL approaches are characterized by how map $\mathcal{M}$, observation $\mathcal{Z}$, and log-likelihood function $\log p(\mathcal{Z} \mid \bm{x}, \mathcal{M})$ are defined. For example, a conventional method using LiDAR scans~\cite{fox1999mcl} assumes a binary occupancy grid map for $\mathcal{M}$ and the set of distances from scan endpoints to their nearest obstacles for $\mathcal{Z}$. The log-likelihood function is given by the sum of terms contributed by scan endpoints. Each term reflects the consistency of the endpoint with the map and assigns higher weights to pose hypotheses with better scan–map match.

\section{VLG-Loc on a Labeled Footprint Map}
We present the VLG-Loc algorithm, which uses \emph{labeled footprint maps} and performs localization by associating these landmarks between the map and image observations.

\begin{figure*}[t]
\centering
\includegraphics[width=1.0\linewidth]{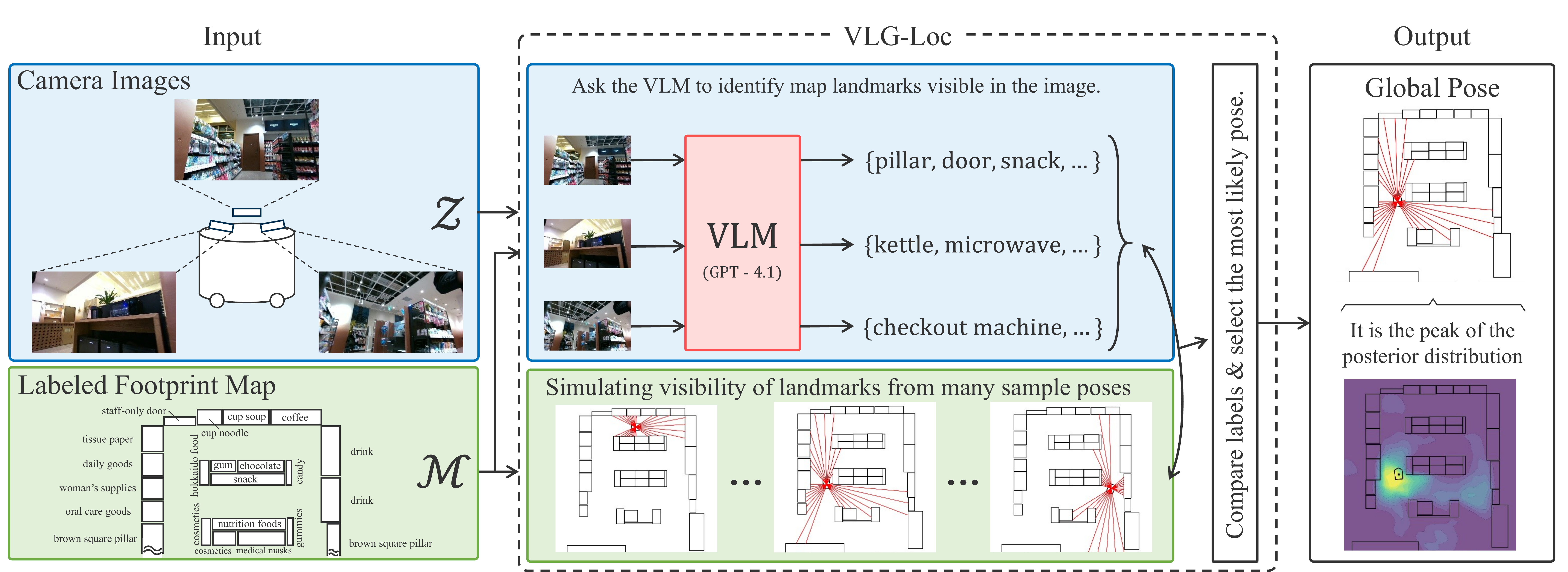}
\caption{
Overview of the proposed VLG-Loc. Given camera images and a labeled footprint map, (i) the vision-language model converts images into landmark labels, and (ii) simulated landmark visibility is computed for hypotheses distributed on the map. The hypothesis best matching the VLM output is taken as the global pose estimate.
}

\label{fig:vlg_loc_overview}
\end{figure*} 

\subsection{Labeled Footprint Maps}
\label{sec:def_of_maps}

A labeled footprint map is a collection of tuples of landmark labels and their footprints:
\begin{align}
    \mathcal{M}:=\left\{(l_i, R_i)\right\},\;l_i \in \mathcal{L},\;R_i\in\mathcal{R}\subset\Omega,
\end{align}
where $\mathcal{L}$ is the set of the text labels of landmarks (\eg, \emph{`door', `table'}) and $\mathcal{R}$ is the set of regions occupied by the landmarks.

This map representation can be regarded as a formal notion of various maps we see every day, such as those in retail stores, offices, or museums. While these maps retain topological consistency, they often omit geometric/appearance details. This simplified form makes it easy for stakeholders (\eg, facility admin or robot operators) to maintain these maps upon environmental changes. For example, even if the contents of the store shelves change, one just needs to rewrite the corresponding text labels, while updating point cloud maps for such changes will require additional data collection and careful stitching of the updated part.

\begin{figure}[t]
  \centering
  \begin{minipage}[t]{0.45\linewidth}
    \centering
    \includegraphics[width=0.75\linewidth]{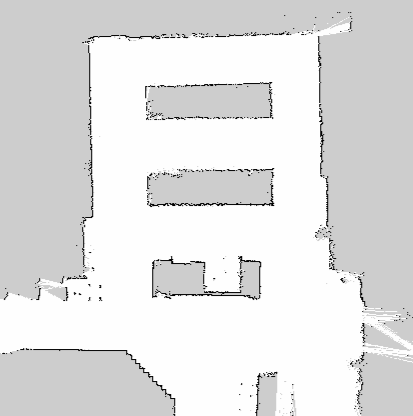}
    \subcaption{$\mathcal{M}_s$: Occupancy Grid Map}\label{fig:left}
    \label{fig:ex_occupancy_grid_map}
  \end{minipage}
  \hfill
  \begin{minipage}[t]{0.53\linewidth}
    \centering
    \includegraphics[width=\linewidth]{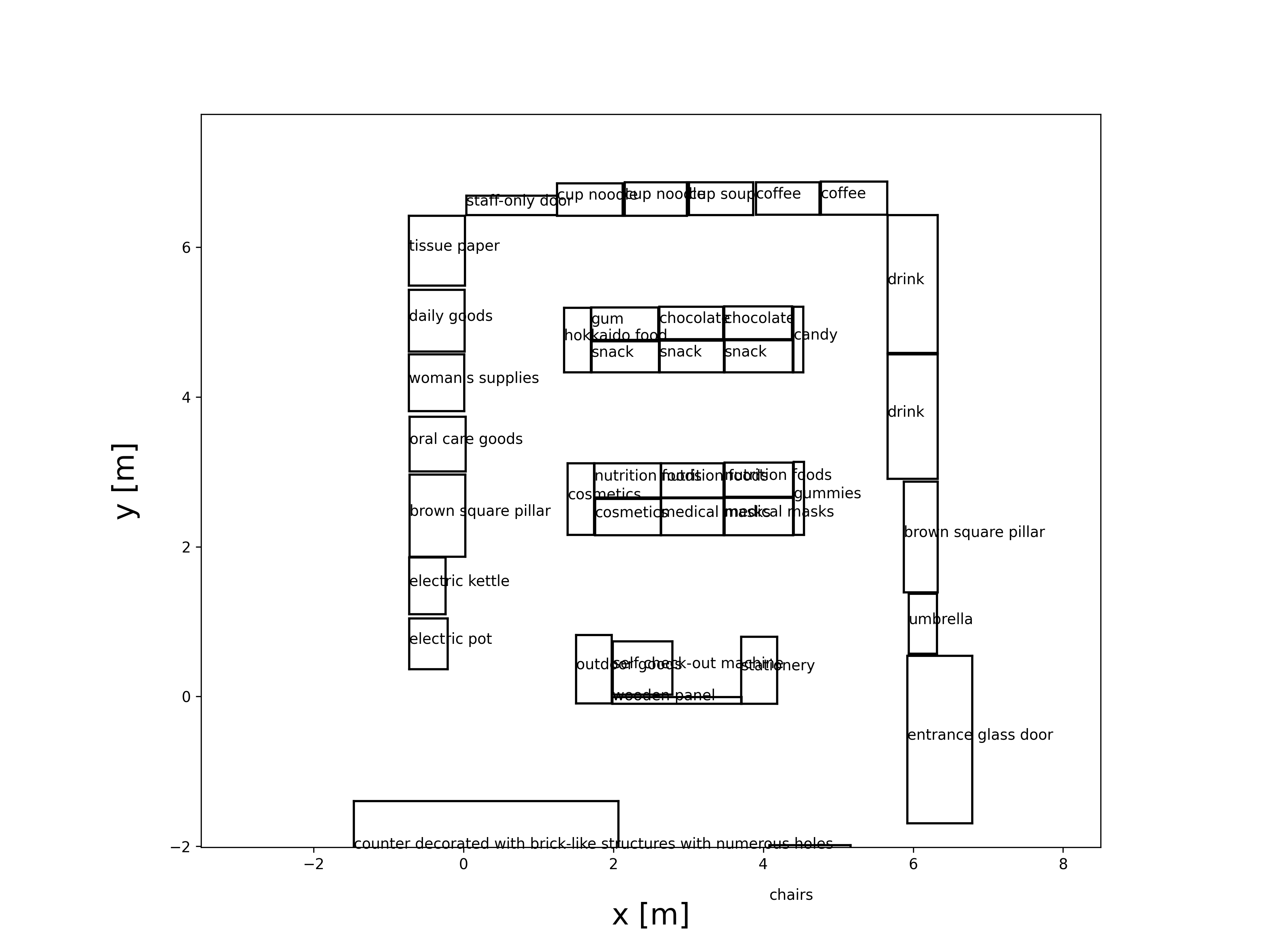}
    \subcaption{$\mathcal{M}$: Labeled Footprint Map}\label{fig:right}
    \label{fig:ex_landmark_map}
  \end{minipage}
  \caption{Two types of maps considered in this study.} 
  \label{fig:two_types_of_maps}
\end{figure}

\subsection{Global Localization with Labeled Footprint Maps}
\label{sec:method_vision}
As illustrated in Fig.~\ref{fig:vlg_loc_overview}, VLG-Loc assumes a mobile robot to mount a set of fixed cameras with known intrinsic and extrinsic parameters (with respect to the robot's base link). Observed images are first analyzed using vision-language models (VLMs) to identify landmarks visible from the robot. Pose hypotheses are then evaluated in the MCL framework based on the consistency of landmark visibility between the image observations and the labeled footprint map.

\looseness=-1
\noindent\textbf{Finding Visible Landmarks in Images.} We ask a VLM to identify landmarks from the list ($\mathcal{L}$ in the previous section) which are clearly visible in each observed image. 
Formally, let $L_i = \{l_0, \dots, l_M\} \subset \mathcal{L},\;i=1,\dots, N$ be \emph{visible labels}, \ie, the list of $M$ landmark labels found in the $i$-th image out of $N$. Note that $L_i$ can be empty $L_i=\emptyset$ when no landmarks are found.

This approach is more flexible than classical object detection techniques~\cite{object_detection_survey_2023} or lightweight language-image encoders (\eg, CLIP-based model~\cite{radford2021learning}), because landmark labels do not require as predefined object classes; they can be fine-grained object categories (\eg, \mytag{snack shelf} rather than \mytag{shelf}) or even proper nouns (\eg, shop names or product brand names) where fine-tuning object detetors for every such landmark is not realistic. 
A significant advantage of using a VLM is its ability to interpret such flexible label representations, enabling the use of human-created maps. This contrasts with previous object-centric approaches~\cite{clip_clique, paul2025sparseloc} that are limited to predefined categories.

\noindent\textbf{Simulating Visible Landmarks.} We can also \emph{simulate} landmarks that should be visible for a given pose, because intrinsic and extrinsic parameters of the cameras are both known. For each camera, we cast virtual rays within the field of view from the pose. If these rays intersect the footprint $R_i$, the landmark $l_i$ should be visible. Let $\bar{L}_i(\bm{x})\subset\mathcal{L}$ be \emph{simulated labels} of visible landmarks simulated in this way for the $i$-th camera from pose $\bm{x}$.

\noindent\textbf{Evaluating Pose Likelihoods.} By regarding visible labels as observation, now we can evaluate the likelihood of pose $\bm{x}$ based on the consistency between visible labels $L_i$ and simulated ones $\bar{L}_i(\bm{x})$ for $N$ cameras. Specifically, we define the consistency score for the $i$-th camera by the cardinality of the intersection set of $L_i$ and $\bar{L}_i(\bm{x})$:
\begin{align}
    S_i(\bm{x})=\left|L_i \cap \bar{L}_i(\bm{x})\right|,
\end{align}
and the score for all the cameras by their sum:
\begin{align}
    S(\bm{x})=\sum_{i=1}^N S_i(\bm{x}).
\end{align}
The pose likelihood is then defined by taking its sigmoid:
\begin{align}
p(\mathcal{Z} \mid \bm{x}, \mathcal{M}):=\sigma(\alpha (S(\bm{x}) - \mu_S)),
\label{eq:vision_normalization_sigmoid}
\end{align}
where $\sigma(t)=(1+\exp^{-t})^{-1}$ is the logistic function, $\alpha>0$ is a scaling factor, and the $\mu_S$ is the mean of the aggregated scores $S(\bm{x})$ over the entire set of hypotheses.
This transformation normalizes the consistency score into $[0, 1]$.
When multiple poses $\bm{x}$ achieve the same maximum likelihood, their average is taken as the maximum a posteriori estimate in order to obtain a unique solution. 

\subsection{Multimodal Extensions to VLG-Loc}
\label{sec:method_fusion}

Practical mobile robot navigation systems employ multimodal sensor fusion techniques to enable robust localization. The proposed VLG-Loc can also be integrated into such systems by combining its vision-based likelihood with that of scan-based localization when available. Scan-based localization provides highly accurate pose estimates when the scan matches uniquely with the environment. However, it may fail severely if the environment contains repetitive or locally similar geometric structures. VLG-Loc, on the other hand, is more robust against such geometric ambiguities by leveraging semantic information from vision. Yet, landmark objects with non-unique or plain appearances may lead to incorrect pose estimates.

We fuse these two complimentary modalities in the MCL framework as follows. Let $\mathcal{M}_\textrm{s}$ be an occupancy grid map defined in the same coordinate system as that of the labeled footprint map $\mathcal{M}$. We also assume the robot to be equipped with a 2D LiDAR sensor, which provides a scan observation $\mathcal{Z}_\textrm{s}$. The likelihood of scan-based localization $p(\mathcal{Z}_\textrm{s} \mid \bm{x}, \mathcal{M}_\textrm{s})$ can be defined as in Sec.~\ref{sec:loc_formulation}, \ie, the distance between the scan endpoints and the nearest obstacles in the occupancy map. We define the fused likelihood by the product of the scan and VLG-Loc likelihoods:
\begin{align}
p(\mathcal{Z}, \mathcal{Z}_\textrm{s} \mid \bm{x}, \mathcal{M}, \mathcal{M}_\textrm{s}) := p(\mathcal{Z} \mid \bm{x}, \mathcal{M}) \cdot (p(\mathcal{Z}_\textrm{s} \mid \bm{x}, \mathcal{M}_\textrm{s}))^{1/\lambda},
\label{eq:fusion_posterior}
\end{align}
where $\lambda > 0$ is a parameter that balances the relative influence of the scan modality. Note that here we assume the conditional independence between the scan and vision modalities. The maximum likelihood solution is finally calculated as follows:
\begin{align}
\bm{x}^* = \underset{\bm{x}}{\arg\max}\;
\Big( \log p(\mathcal{Z} \mid \bm{x}, \mathcal{M})
+ (1 / \lambda) \log p(\mathcal{Z}_\textrm{s} \mid \bm{x}, \mathcal{M}_\textrm{s}) \Big).
\label{eq:fusion_map}
\end{align}

\section{Experiments}

We systematically evaluate the effectiveness of the proposed method over several scan-based localization baselines~\cite{fox1999mcl,cbgl2024iros} in both simulation and real-world environments.

\subsection{Environments}

\paragraph{Robot Setup}
We constructed a mobile robot platform equipped with an autonomous mobile robot (Preferred Robotics Kachaka), and three cameras (Intel RealSense Depth Camera D455) mounted horizontally on top of the robot with their optical axes separated by 120$^\circ$ (see also Fig.~\ref{fig:teaser}.) 
The mobile robot internally includes an IMU, a 2D LiDAR, and wheel odometry for scan-based localization.
In what follows, we adopted this platform for data collection from the real-world environment, and its simulated version for the simulation environments.

\paragraph{Simualtion Environments}
We systematically designed four simulation environments that are either uniform or diverse in their geometric layout and/or appearance as shown in Tab.~\ref{tab:perf_methods_sim_env}, which we called UG/UA (Uniform Geometry/Uniform Appearance), UG/DA (Uniform Geometry/Diverse Appearance), DG/UA (Diverse Geometry/Uniform Appearance), and DG/DA (Diverse Geometry/Diverse Appearance) environments. Comparing localization performances across these environments will allow us to investigate how each method relies on geometric and/or visual cues.
\begin{itemize}
\item \textbf{UG/UA} is an environment of identical columns arranged in a regular pattern. In this case, neither geometric nor visual information provides useful cues for localization.
\item \textbf{UG/DA} consists of regularly placed bookshelves. While geometric cues remain limited, each shelf carries a distinct alphabetical label providing rich visual information.
\item \textbf{DG/UA} represents an indoor scene with many pieces of furniture. Geometric cues are abundant due to the dense distribution of objects. However, repeated items make visual localization challenging. 
\item \textbf{DG/DA} is also an indoor environment with numerous objects. In this case, the greater variety of furniture reduces repetition. Landmarks can be more reliably linked to positions than in the DG/UA setting.
\end{itemize}
Each environment is associated with an occupancy grid map and a labeled footprint map, as shown in Fig.~\ref{fig:two_types_of_maps}.
Occupancy maps are extracted directly from the environment with gazebo\_map\_creator\footnote{\url{https://github.com/arshadlab/gazebo_map_creator}}, providing the ground-truth geometric information.  
The object names are manually annotated to the occupancy maps to construct labeled footprint maps in a consistent coordinate system.
Although the labeling process was not highly precise, this design choice was intended to reflect realistic conditions in which semantic annotations are often inaccurate or incomplete.

\paragraph{Retail Environment}
To examine the applicability to real-world scenarios, we evaluated localization performance using data collected in an actual retail store of approximately 70 m$^2$ during regular operation as shown in Tab.~\ref{tab:perf_methods_real_env}.
{We first generated the occupancy grid map using SLAM functionality of the mobile platform, and then annotated shelf labels (\eg, \mytag{snack} and \mytag{drink}) with reference to a product layout map provided by the retailer. 
We also added some additional labels of objects which may be useful for landmarks, such as a \mytag{microwave oven} and a \mytag{staff-only door}.}

\paragraph{Retail Environment with Substituted Appearances}
As also shown in Tab.~\ref{tab:perf_methods_real_env}, we additionally created an environment that replaced the appearance of the retail environment with tiles with non-overlapping alphanumeric characters. While retaining the original geometric layout, the labeled footprint map was also replaced by those characters accordingly. Comparing results between the original retail environment and this appearance-substituted version will reveal how our approach is affected by real-world items with similar appearances but different labels.

\subsection{Evaluation Setup}

\paragraph{VLG-Loc Implementation}
In the image recognition pipeline of VLG-Loc, we employed the GPT-4.1 model as a VLM, with the following prompt template:
\begin{quote}
\small\texttt{You are an image recognition assistant. From the list below,
identify only the objects that are clearly visible in the image. Include
partially visible objects. Do not include any object if you are not
confident it is present. Object list: [OBJECT LIST]}
\end{quote}
where \texttt{[OBJECT LIST]} was the list of landmark names extracted from a given labeled footprint map.

To simulate which landmarks are visible from the three cameras for a given pose, we approximated each camera's field of view by the 10 rays in a ray-casting process, as shown in Fig.~\ref{fig:vlg_loc_overview}.
In the multimodal fusion process, we empirically set the sigmoid parameter $\alpha = 0.5$ in Eq.~\ref{eq:vision_normalization_sigmoid}, and the temperature parameter to $\lambda=1500.0$
\footnote{
This large value of λ serves to balance the influence of the two sensor modalities.
The scan-based likelihood distribution exhibits sharper peaks, resulting in a much larger range of log-likelihood values, compared to the broader vision-based distribution. 
}
in Eq.~\ref{eq:fusion_posterior}.

\paragraph{Baseline Methods}
As a baseline, we evaluated a conventional scan-based Monte Carlo global localization~\cite{fox1999mcl} as well as the recent method called CBGL that combines Monte Carlo sampling and scan-to-map registration~\cite{cbgl2024iros}. 
For Monte Carlo sampling, one million particles (\ie, pose hypotheses) were sampled uniformly from the entire map with fixed random seeds. 
The sample poses are exactly the same as the proposed VLG-Loc to ensure the fairness and reproducibility.
For CBGL, we used the default parameters of its publicly available implementation.

\paragraph{Dataset Construction}
For each environment, we constructed an evaluation dataset by recording the sensing information (\ie, 2D laser scan and camera images) of a robot navigating through the environment 
at intervals of 1--2 seconds, adjusted according to the environment.
The ground-truth poses were estimated using a particle filter-based localization method~\cite{emcl2} that fully utilized occupancy grid maps, laser scan measurements, and odometry data.

\paragraph{Evaluation Metrics}
We employed the translational error and the rotational error as evaluation metrics.
Specifically, let the ground-truth pose be $\bm{x}_{\mathrm{gt}}=(x_{\mathrm{gt}}, y_{\mathrm{gt}}, \theta_{\mathrm{gt}})$ and the estimated pose be $\bm{x}_{\mathrm{est}}=(x_{\mathrm{est}}, y_{\mathrm{est}}, \theta_{\mathrm{est}})$. 
The translational error is defined as the Euclidean distance between the ground-truth and estimated positions,
$e_{\mathrm{trans}} = \sqrt{(x_{\mathrm{est}}-x_{\mathrm{gt}})^2 + (y_{\mathrm{est}}-y_{\mathrm{gt}})^2}$. 
The rotational error is defined as the absolute angular difference between the estimated and ground-truth orientations, normalized to the interval $[-\pi, \pi)$,
$e_{\mathrm{rot}} = \left|\mathrm{wrap}(\theta_{\mathrm{est}} - \theta_{\mathrm{gt}})\right|,$
where $\mathrm{wrap}(\cdot)$ denotes the angle wrapping function
$\mathrm{wrap}(\phi) = \arctan2(\sin \phi, \cos \phi)$.
This formulation ensures the angular difference to take the minimum direction within the range $[0, \pi]$. 
\subsection{Experimental Results}

\subsubsection{Comparison in Simulation Environments}

\begin{table*}[t]
  \vspace{5pt} 
  \centering
  \begin{threeparttable}
  \caption{Global localization accuracy in simulated environments categorized by geometric and visual characteristics.} 
  \label{tab:perf_methods_sim_env}

\begin{tabular}{c|cccc}
    \toprule
    \multirow[c]{8}{*}{\textbf{Localization Method}} 
      & \textbf{UG/UA Env.} & \textbf{UG/DA Env.} 
      & \textbf{DG/UA Env.} & \textbf{DG/DA Env.} \\
    
    & \includegraphics[width=0.12\textwidth]{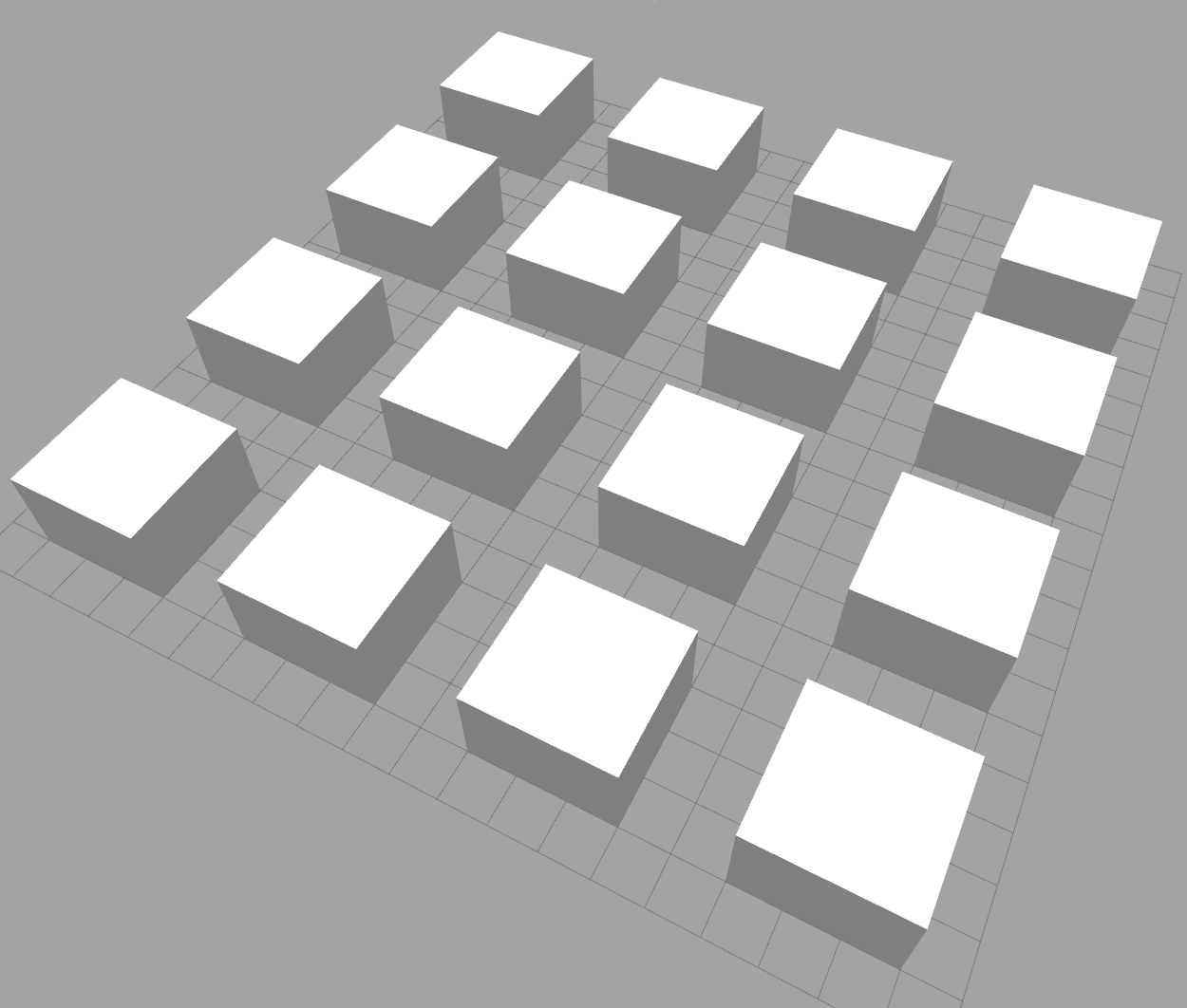}
    & \includegraphics[width=0.12\textwidth]{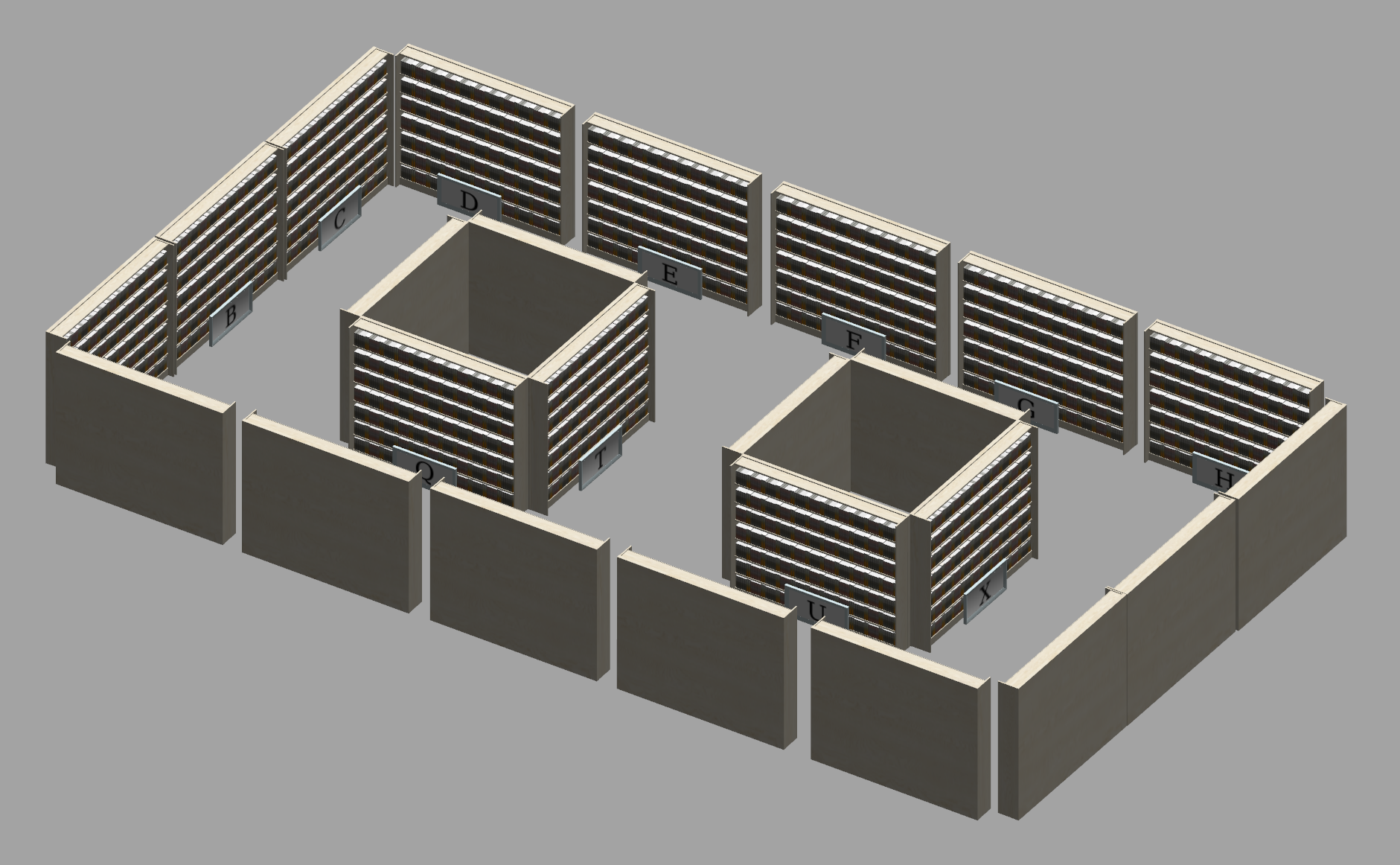}
    & \includegraphics[width=0.12\textwidth]{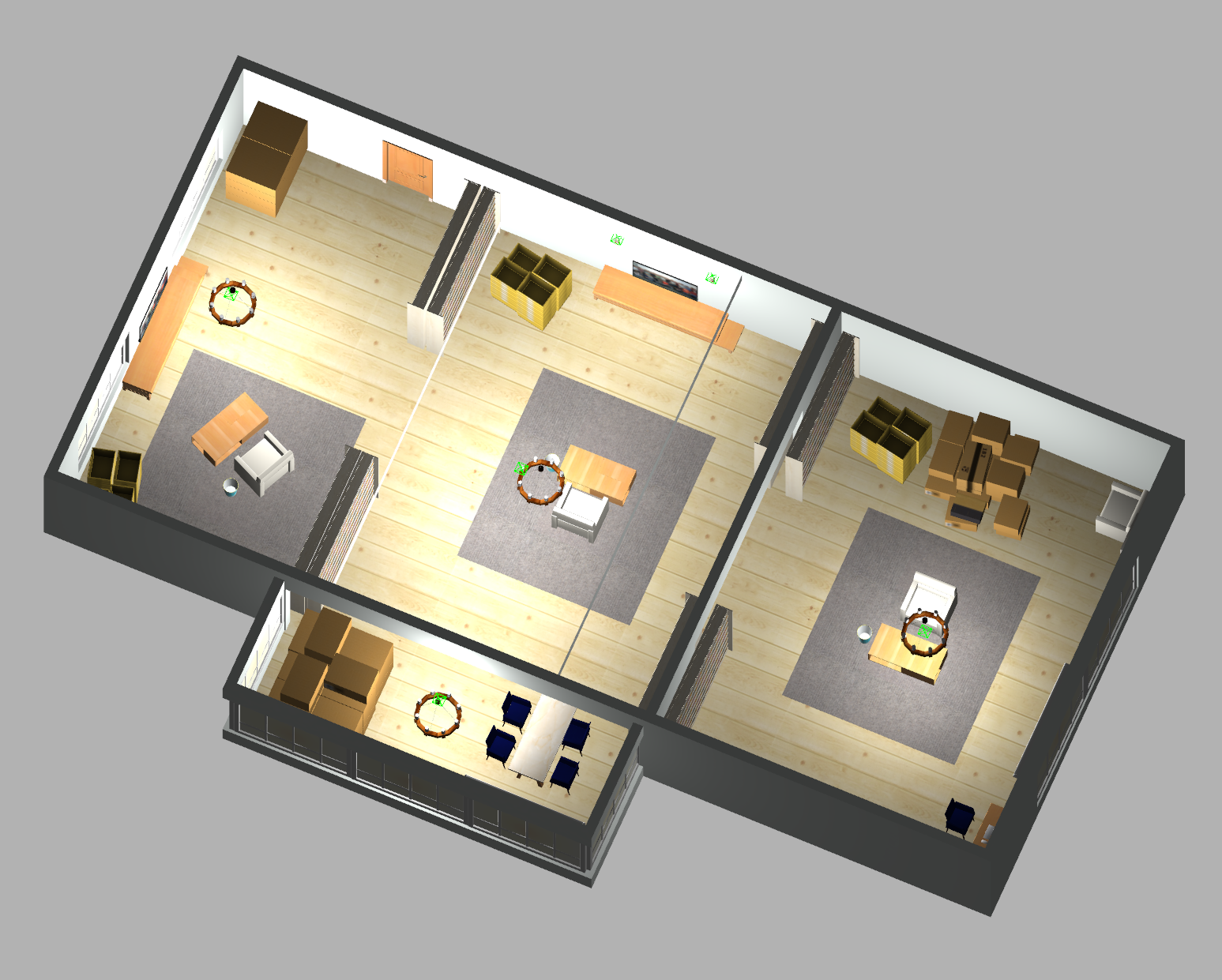}
    & \includegraphics[width=0.12\textwidth]{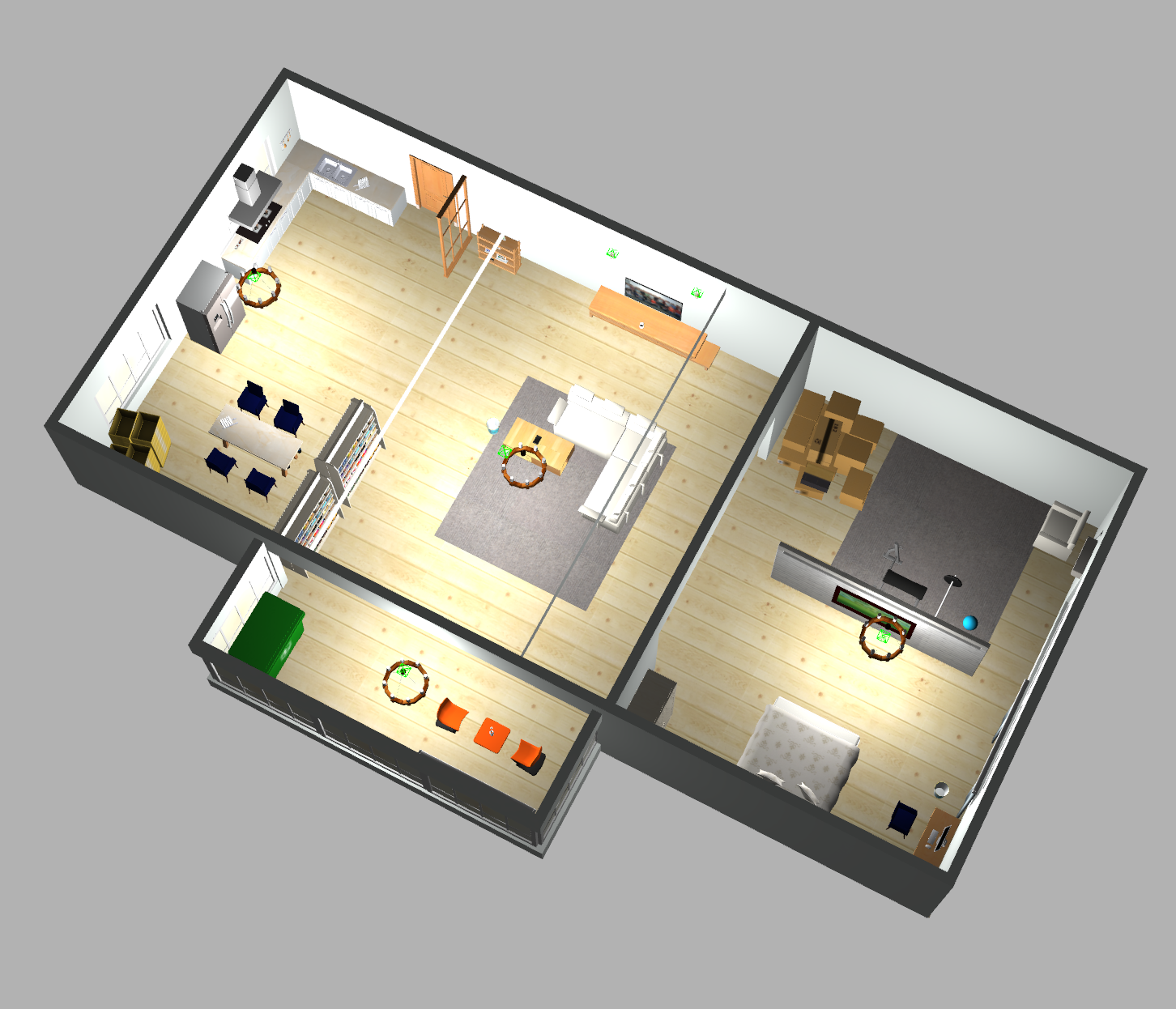} \\
    \midrule

    Scan-Based Localizer (MCL) &
      $6.94\!\pm\!3.71\ /\ 1.82\!\pm\!1.05$ &
      $3.39\!\pm\!4.01\ /\ 1.29\!\pm\!1.48$ &
      $2.15\!\pm\!3.49\ /\ 0.62\!\pm\!1.04$ &
      $2.26\!\pm\!3.89\ /\ 0.61\!\pm\!1.05$ \\

    Scan-Based Localizer (CBGL) &
      $\mathbf{6.90}\!\pm\!3.99\ /\ \mathbf{1.69}\!\pm\!1.10$ &
      $3.62\!\pm\!4.10\ /\ 1.17\!\pm\!1.25$ &
      $1.68\!\pm\!2.91\ /\ 0.70\!\pm\!1.01$ &
      $2.11\!\pm\!3.43\ /\ 0.64\!\pm\!0.89$ \\ 

    VLG-Loc (Vision Only) &
      $11.62\!\pm\!3.89\ /\ 1.78\!\pm\!0.76$ &
      $1.08\!\pm\!0.62\ /\ 0.24\!\pm\!0.14$ &
      $2.36\!\pm\!2.22\ /\ 0.63\!\pm\!0.76$ &
      $1.65\!\pm\!1.96\ /\ 0.33\!\pm\!0.51$ \\

    VLG-Loc (Vision and Scan) &
      $9.39\!\pm\!3.53\ /\ 1.82\!\pm\!1.05$ &
      $\mathbf{0.21}\!\pm\!0.28\ /\ \mathbf{0.03}\!\pm\!0.03$ &
      $\mathbf{0.67}\!\pm\!1.69\ /\ \mathbf{0.21}\!\pm\!0.72$ &
      $\mathbf{0.62}\!\pm\!1.43\ /\ \mathbf{0.12}\!\pm\!0.44$ \\
    \bottomrule
  \end{tabular}
  \begin{tablenotes}
    \footnotesize
    \item Values show \textbf{Trans./Rot.} error (mean $\pm$ std). 
  Best per metric in bold.
  \end{tablenotes}
  \end{threeparttable}
\vspace{-3mm}
\end{table*}

\begin{table}[t]
  \centering
  \begin{threeparttable}
  \caption{Evaluation results in the retail environments. }
  \label{tab:perf_methods_real_env}
  \begin{tabular}{c|cc}
    \toprule
      &
      \textbf{\makecell{Retail Env.}} &
      \textbf{\makecell{Retail Env.\\(Subst.\ Appear.)}} \\
    \midrule
        \textbf{\makecell{Method}} &
        \includegraphics[width=0.12\textwidth]{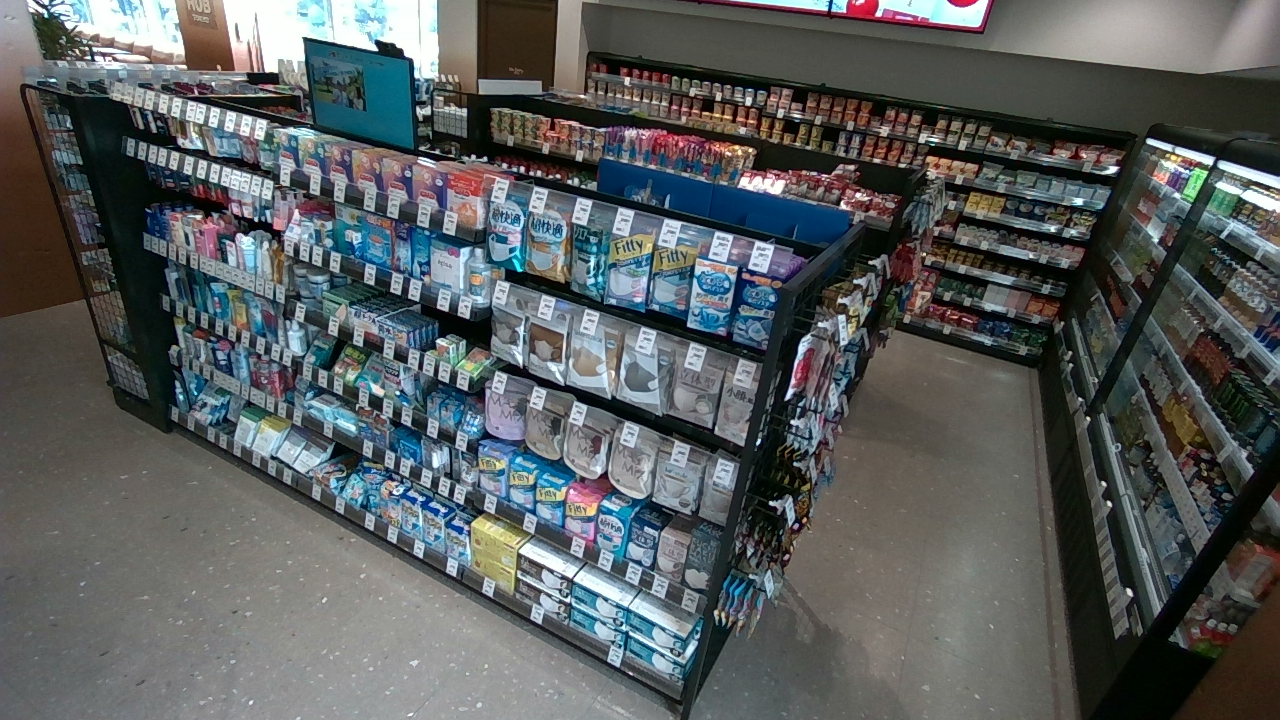} &
        \includegraphics[width=0.12\textwidth]{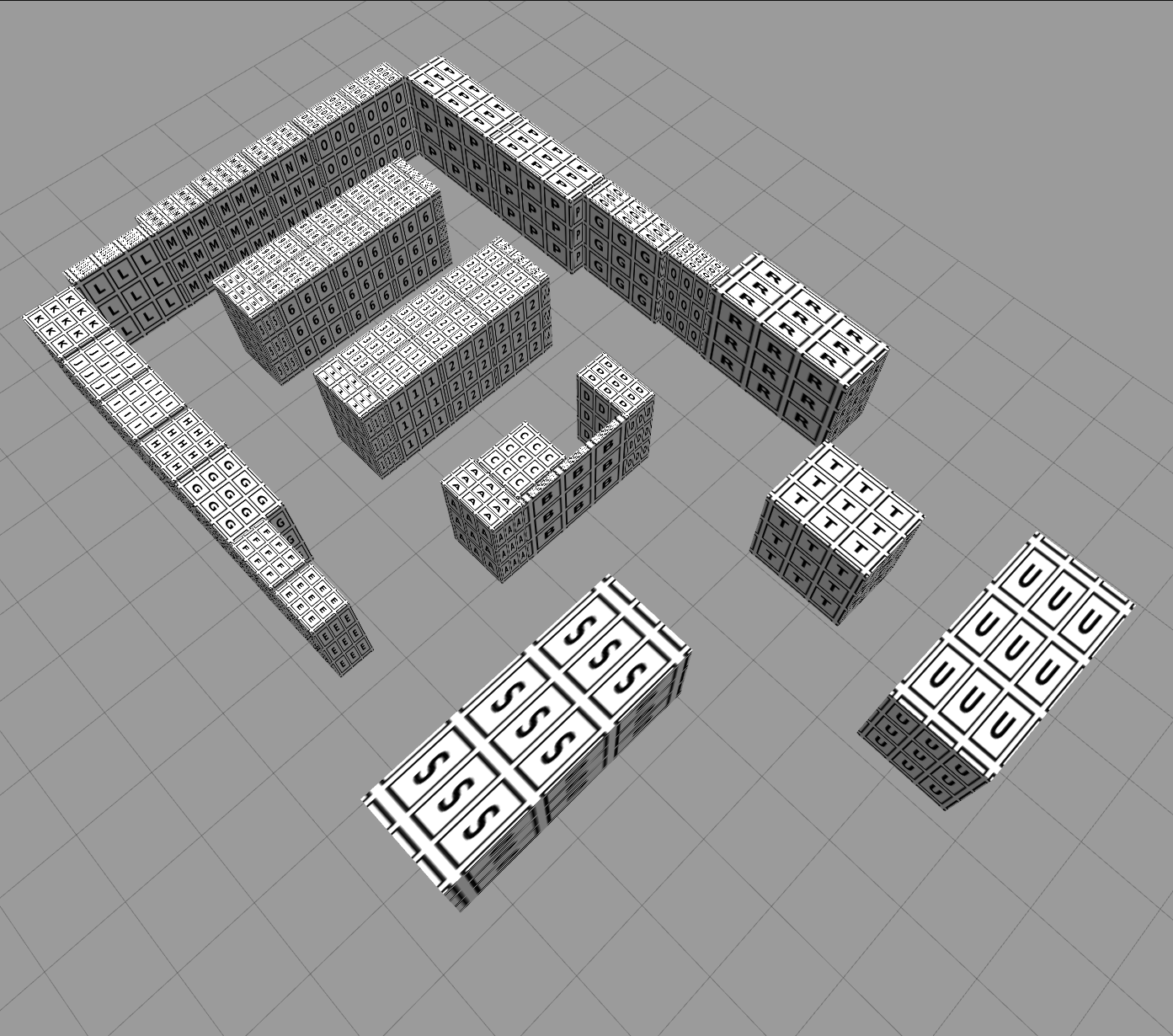} \\
    \midrule

    \makecell{MCL} &
      $1.97\!\pm\!2.05\ /\ 1.01\!\pm\!1.11$ &
      $0.69\!\pm\!1.11\ /\ 0.42\!\pm\!0.94$ \\

    \makecell{CBGL} &
      $2.31\!\pm\!1.98\ /\ 1.26\!\pm\!1.14$ &
      $1.97\!\pm\!1.81\ /\ 1.02\!\pm\!1.06$ \\

    \makecell{VLG-Loc (VO)} &
      $1.08\!\pm\!0.78\ /\ 0.46\!\pm\!0.49$ &
      $0.43\!\pm\!0.23\ /\ 0.17\!\pm\!0.16$ \\

    \makecell{VLG-Loc (VS)} &
      $\mathbf{0.52}\!\pm\!0.92\ /\ \mathbf{0.19}\!\pm\!0.52$ &
      $\mathbf{0.18}\!\pm\!0.41\ /\ \mathbf{0.09}\!\pm\!0.42$ \\
    \bottomrule
  \end{tabular}

  \begin{tablenotes}
    \footnotesize
    \item VO and VS mean Vision Only and Vision and Scan, respectively. Values show \textbf{Trans./Rot.} error (mean $\pm$ std). 
  \end{tablenotes}
  \end{threeparttable}
\end{table}

Table~\ref{tab:perf_methods_sim_env} lists the evaluation results for the simulation environments. Overall, the diversity and distinctiveness of landmarks are critical factors in uniquely identifying poses, making the UG and UA environments (\ie, uniform geometry and uniform appearance) the most challenging to solve. Specifically, scan-based approaches (MCL and CBGL) struggle in UG environments due to the lack of geometric cues, whereas our proposed method, VLG-Loc (Vision Only), excels in DA environments by leveraging rich visual cues. On the other hand, VLG-Loc (Vision Only) faces difficulties in UA environments because of the scarcity of distinctive visual cues. 
These trends highlight the importance of cross-modal fusion, especially in visually or geometrically repetitive environments.

\looseness=-1 
Figure~\ref{fig:results_examples_overview} (a) and (b) illustrate typical examples of estimation results in the UG/DA and DG/UA environments.
As revealed in these examples, repeated or uniform cues lead to ambiguous, multi-modal likelihood distributions that degrade localization accuracy.
In the (a) UG/DA environment, similar geometric structures produce multiple plausible scan matches, while in (b) DG/UA, visually repetitive patterns confuse vision-based estimation.
Such ambiguities increase translational errors and reduce pose stability as shown in Tab.~\ref{tab:perf_methods_sim_env}.
Conversely, distinctive geometric or visual cues yield unimodal likelihood distributions and more consistent localization results.

\looseness=-1 
The proposed VLG-Loc (Vision and Scan) effectively complements the weaknesses of each modality.
As shown in Fig.~\ref{fig:results_examples_overview}, relying solely on one modality often leads to multi-modal likelihoods, whereas combining both modalities mitigates this issue, enabling more accurate estimation.
Indeed, as seen in Tab.~\ref{tab:perf_methods_sim_env}, VLG-Loc (Vision and Scan) achieves the highest accuracy in all environments except UG/UA.
Note that while UG/UA is an exception due to the scarcity of visual cues that make vision-based localization nearly infeasible, such environments are rare in real-world scenarios.

\subsubsection{Comparison in a Real-world Retail Environment}
\label{sec:exp_real_env}

We also evaluated our method in real-world retail environments, as summarized in Tab.~\ref{tab:perf_methods_real_env}.
These environments are similar to the UG/DA setting in simulation, where VLG-Loc (Vision and Scan) achieved the highest accuracy.
In particular, because of the presence of visually diverse landmarks, VLG-Loc (Vision Only) achieved better localization performance than scan-based methods. However, unlike in simulation, misrecognition by the VLM relatively limits vision-based localization performance.
The Substituted Appearance (Subst. Appear.) environment mimics the layout and object placement of the retail environment but uses easily distinguishable object labels.
Consequently, as shown in Tab.~\ref{tab:perf_methods_real_env}, VLG-Loc achieved higher performance in the Subst. Appear. environment than in the real retail environment.

\subsection{Failure Cases and Limitations}
\begin{figure}[t]
  \centering
  \begin{minipage}[t]{0.32\linewidth}
    \centering
    \includegraphics[width=\linewidth]{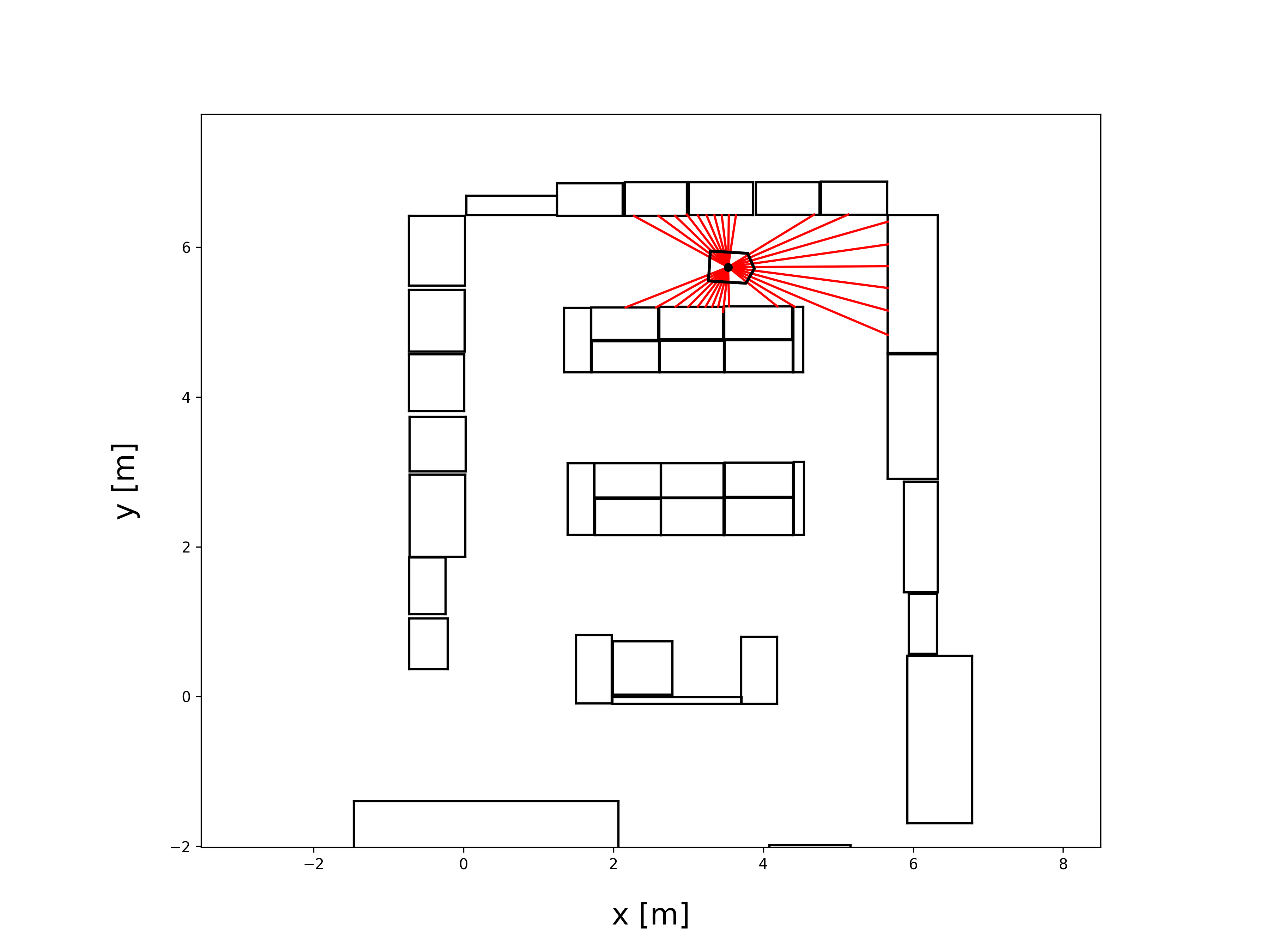}
    \subcaption{Ground Truth}\label{fig:gt}
    \label{fig:ex_discussion_no_snack_map}
  \end{minipage}  
  \begin{minipage}[t]{0.32\linewidth}
    \centering
    \includegraphics[width=\linewidth]{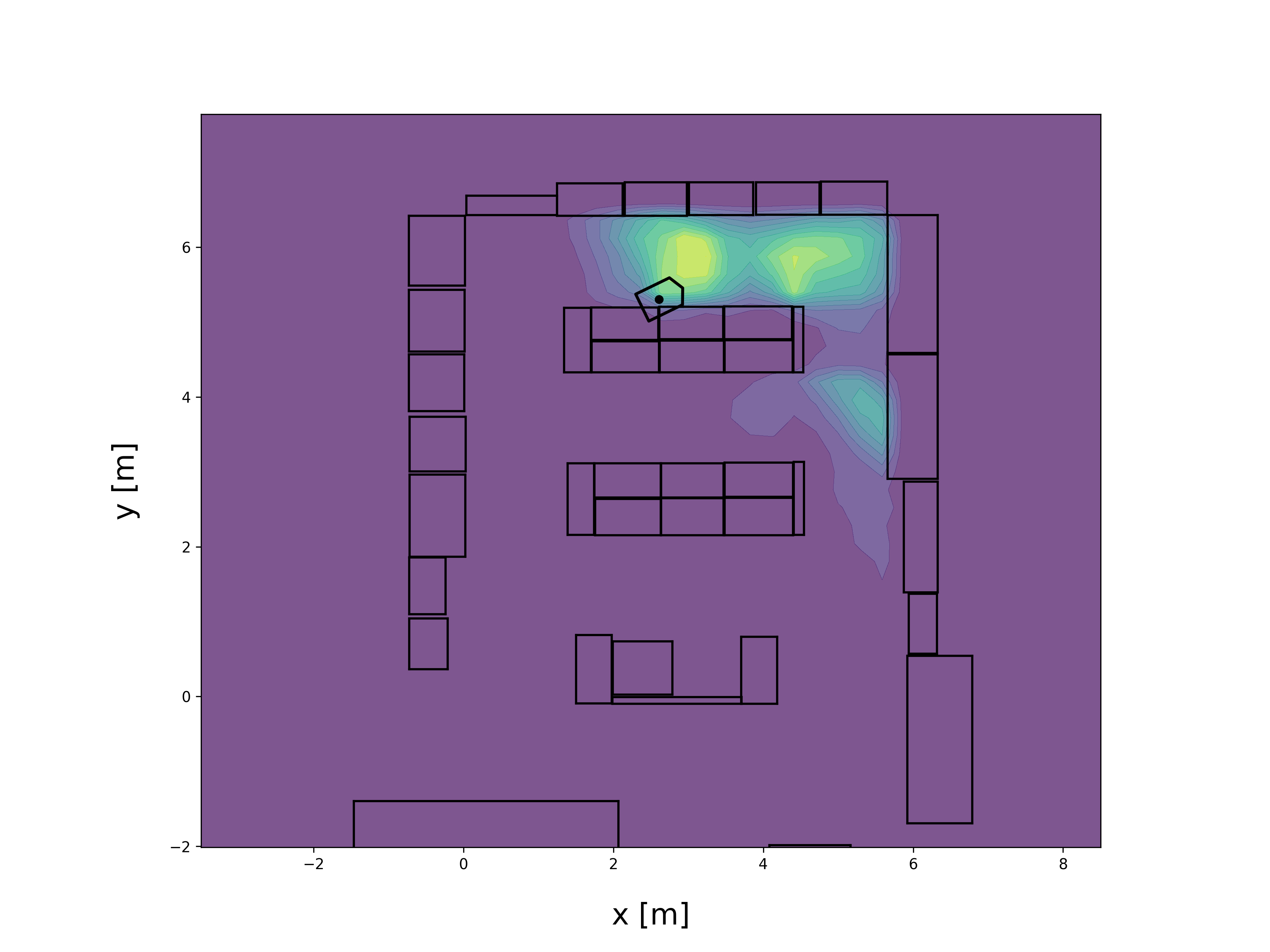}
    \subcaption{With \mytag{snack}}\label{fig:left}
    \label{fig:ex_discussioin_complete_map}
  \end{minipage}
  \hfill
  \begin{minipage}[t]{0.32\linewidth}
    \centering
    \includegraphics[width=\linewidth]{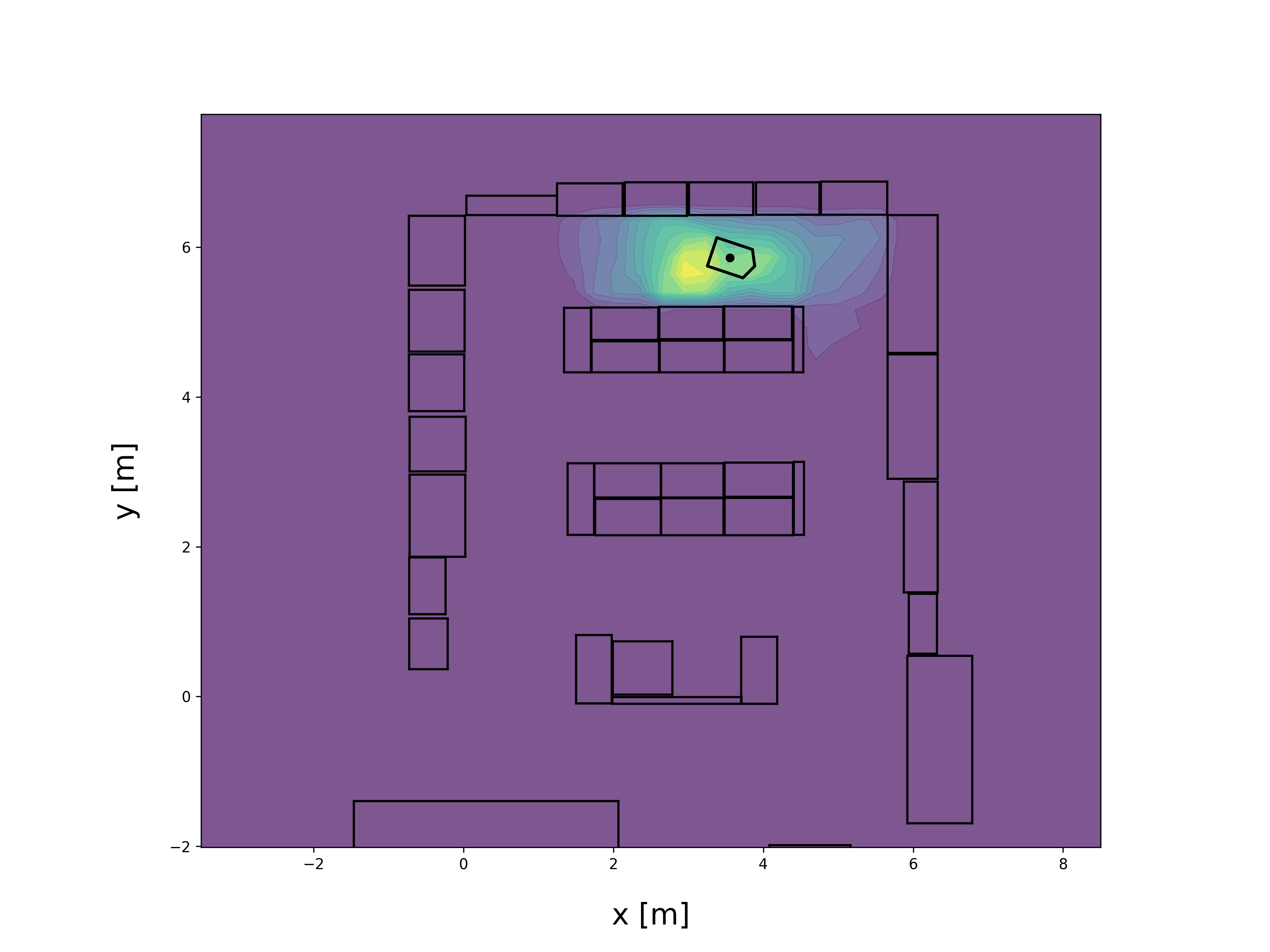}
    \subcaption{Without \mytag{snack}}\label{fig:right}
    \label{fig:ex_discussion_no_snack_map}
  \end{minipage}
  \caption{Effect of excluding the \mytag{snack} label from the labeled footprint map. By excluding the \mytag{snack} label, the multimodal estimated likelihood distribution becomes unimodal, resulting in an estimation closer to the ground truth.}
  \label{fig:ex_discussion}
\end{figure}

\begin{figure*}[t]
\centering
\includegraphics[width=1.0\linewidth]{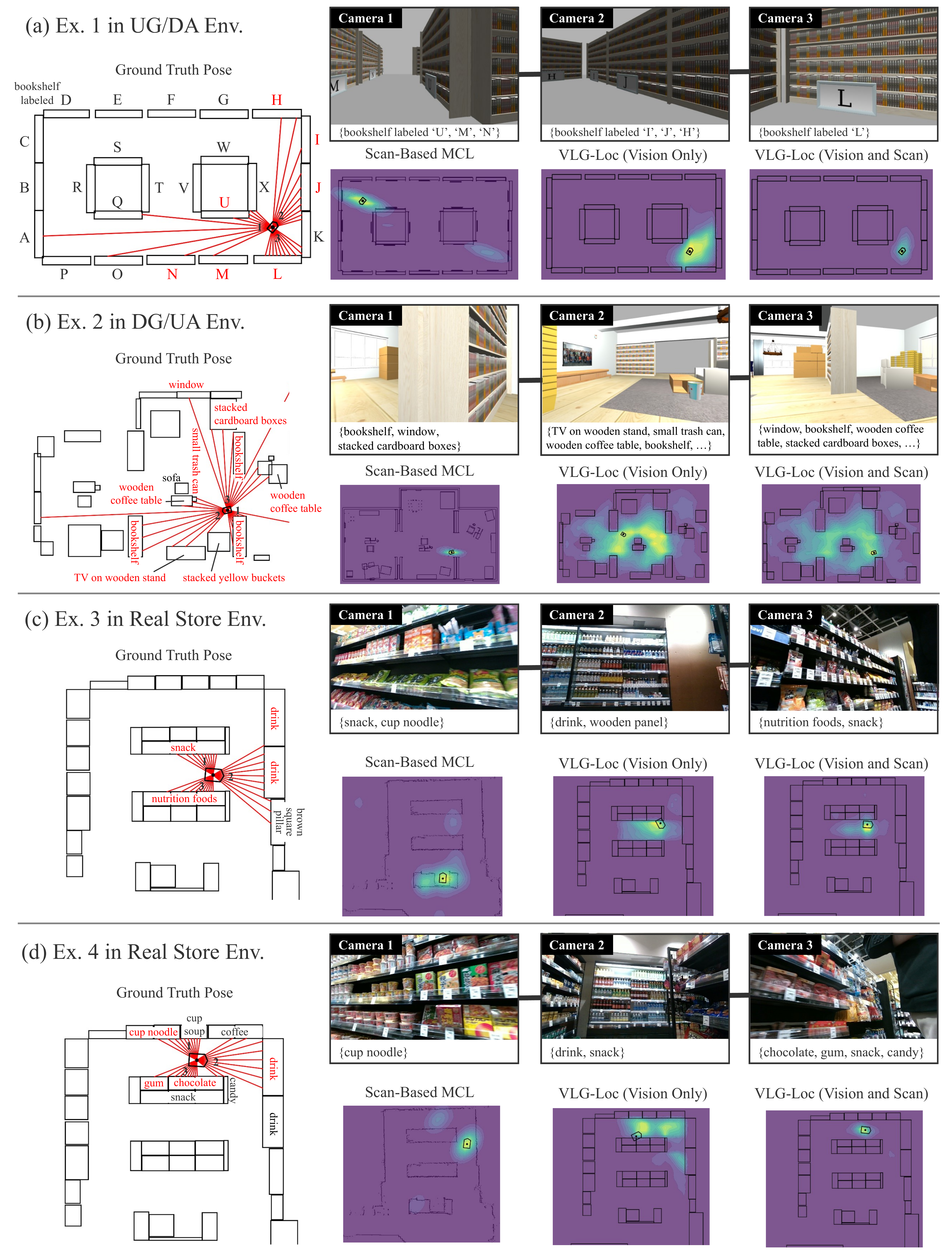}
\caption{
Examples of camera observations and estimation results. 
Red labels indicate correctly recognized objects.
}
\label{fig:results_examples_overview}
\end{figure*}

Figure~\ref{fig:results_examples_overview} (c) shows examples of misrecognition: the VLM recognizes \mytag{cup noodle} and \mytag{wooden panel} that however do not exist in the labeled footprint map.
Nevertheless, the distribution estimated by VLG-Loc (Vision Only) remains unimodal and roughly indicates the correct position.
On the other hand, Fig.~\ref{fig:results_examples_overview} (d) shows that the VLM categorizes \mytag{gum} and \mytag{chocolate} as \mytag{snack}.
As a result, the estimated distribution becomes multimodal, leading to degraded localization accuracy compared to Fig.~\ref{fig:results_examples_overview} (c).
In fact, this causal relationship is confirmed in Fig.~\ref{fig:ex_discussion}, which shows the localization result after excluding the \mytag{snack} category from the map.
The estimated distribution becomes unimodal, and the localization accuracy improves significantly.

These results suggest a trade-off between the abstraction level of landmark labels in the labeled footprint map and localization accuracy.
That is, when the label abstraction is high, such as \mytag{snack}, other related landmarks like \mytag{gum} and \mytag{chocolate} may be erroneously triggered, which can cause incorrect localization.
Ensuring a one-to-one correspondence between labels and their visual appearances is preferable for reliable localization.
However, excessively reducing the level of label abstraction may make recognition by the VLM itself difficult and increase map maintenance costs.

Finally, we identified several limitations of the proposed method and future research directions that remain:
\begin{itemize}
    \item \textbf{Communication latency of VLM inference:}
    Our current implementation relies on an external API service (\ie, GPT-4.1) for VLM inference, which introduces non-negligible communication latency. We chose this approach because our target task, global localization, prioritizes estimation accuracy over inference speed. Nevertheless, this latency presents a potential bottleneck for real-time applications.
    \item \textbf{\emph{Optimal} labeled footprint map generation:}
    Although we employed a manually created labeled footprint map, there exists a trade-off between the abstraction level of landmark labels and localization accuracy as discussed above.
    The automatic optimization of the abstraction level is a promising future direction that can enhance the versatility and scalability of VLG-Loc.
\end{itemize}

\section{Conclusion}
This paper presents Vision-Language Global Localization (VLG-Loc), a method that estimates the global robot pose by integrating camera images with a labeled footprint map.
The VLM plays a crucial role in linking objects observed in camera images to corresponding landmarks annotated on the map.
The proposed approach addresses the limitations of scan-based methods by leveraging abundant visual landmarks represented by text labels in the map, particularly in geometrically uniform environments such as retail stores.
Moreover, the fusion of vision and scan data demonstrated improved localization performance across diverse environments.


\bibliographystyle{IEEEtran}
\bibliography{ref}

\end{document}